\documentclass[lettersize,journal]{IEEEtran}
\usepackage{amsmath,amsfonts}
\usepackage{bm}
\usepackage{algorithm}
\usepackage{algpseudocode}
\usepackage{array}
\usepackage[caption=false,font=normalsize,labelfont=sf,textfont=sf]{subfig}
\usepackage{textcomp}
\usepackage{stfloats}
\usepackage{url}
\usepackage{verbatim}
\usepackage{graphicx}
\def\BibTeX{{\rm B\kern-.05em{\sc i\kern-.025em b}\kern-.08em
    T\kern-.1667em\lower.7ex\hbox{E}\kern-.125emX}}
\usepackage{balance}

\usepackage{hyperref}
\usepackage{multirow}
\usepackage[table]{xcolor}

\usepackage{pifont}
\newcommand{\xmark}{\ding{55}}%
\newcommand{\cmark}{\ding{51}}%

\newcommand{\modified}[1]{{\color{black} #1}}
\newcommand{\updated}[1]{{\color{black} #1}}
\newcommand{\upupdated}[1]{{\color{black} #1}}
\newcommand{\upupupdated}[1]{{\color{black} #1}}
\newcommand{\upupupupdated}[1]{{\color{black} #1}}

\begin{document}

\title{D4D: An RGBD diffusion model to boost monocular depth estimation}

\author{Lorenzo Papa~\IEEEmembership{Student Member IEEE}, Paolo Russo and Irene Amerini~\IEEEmembership{Member IEEE}
\thanks{The authors are with the  Department of Computer, Control and Management Engineering, Sapienza University of Rome, Italy (e-mail: [papa, paolo.russo, amerini]@diag.uniroma1.it).}
\thanks{Manuscript received Month Day, 2022; revised Month Day, 2022.}
}

\markboth{Journal of \LaTeX\ Class Files,~Vol.~18, No.~9, September~2020}%
{How to Use the IEEEtran \LaTeX \ Templates}

\maketitle

\begin{figure*}[t]
    \includegraphics[width=\linewidth]{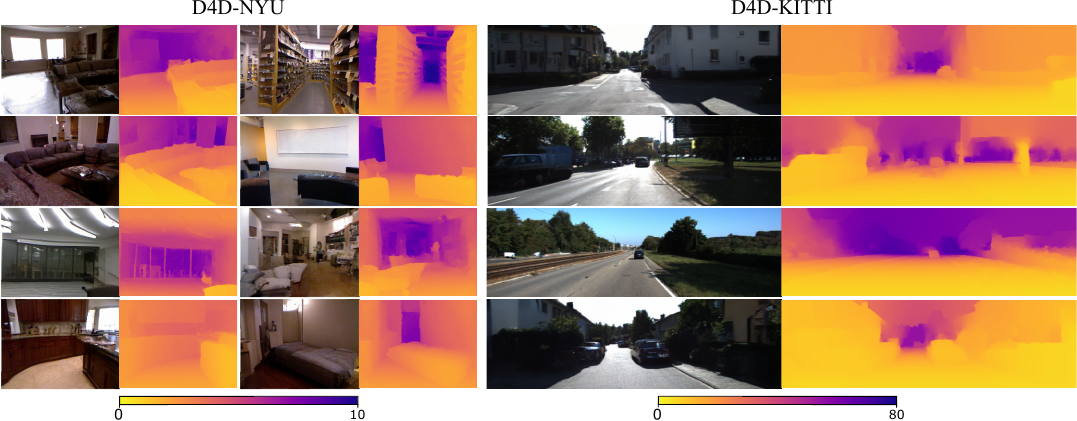}
    \caption{\upupdated{D4D generated RGBD samples based on the indoor NYU Depth v2 (right) and the outdoor KITTI (left) datasets. 
    The images are scaled to match the aspect ratio of the original samples.
    The depth maps are converted in RGB format with a perceptually uniform colormap for a better view, while the two bottom colorbars emphasize the depth data distribution (in meters) over the generated samples.}}
    \label{fig:cover}
\end{figure*}

\begin{abstract}
Ground-truth RGBD data are fundamental for a wide range of computer vision applications; however, those labeled samples are difficult to collect and time-consuming to produce.
A common solution to overcome this lack of data is to employ graphic engines to produce synthetic proxies; however, those data do not often reflect real-world images, resulting in poor performance of the trained models at the inference step.
In this paper we propose a novel training pipeline that incorporates Diffusion4D (D4D), a customized 4-channels diffusion model able to generate realistic RGBD samples.
We show the effectiveness of the developed solution in improving the performances of deep learning models on the monocular depth estimation task, where the correspondence between RGB and depth map is crucial to achieving accurate measurements.
\updated{Our supervised training pipeline, enriched by the generated samples, outperforms synthetic and original data performances achieving an RMSE reduction of 
($8.2\%$, $11.9\%$) and ($8.1\%$, $6.1\%$) respectively on the indoor NYU Depth v2 and the outdoor KITTI dataset.}
\end{abstract}



\section{Introduction}
\label{sec:introduction}
Deep learning has achieved astonishing results in several research fields encouraging its fast growth in all of its aspects, from the study of neural network structure to its optimization.
In computer vision and image processing, it has gained significant success in tasks like object detection, depth estimation, and semantic segmentation~\cite{ioannidou2017deep}.
However, the increasing size and capacity of neural network architectures require the availability of a huge amount of labeled training data, which are often missing or difficult to collect. 
This issue led researchers to focus on several techniques to reduce the data requirements, such as unsupervised \cite{unsup_survey} or self-supervised \cite{jing2020self} learning strategies, with the objective of categorizing unlabeled or partially labeled data.
However, unsupervised learning is intrinsically more complex than (data-driven) supervised learning due to the lack of labeled output samples.
Another possible solution could be the use of AI-based methodologies \cite{generation_survey} to automatically generate realistic samples and data augmentation techniques \cite{data_aug_survey} exploited to increase the diversity of training data.
Nevertheless, the latter techniques are usually constrained by the mathematical transformations that can be used to modify original images while preserving their information.
Moreover, the automatic generation of realistic samples has been typically attributed to variational autoencoders (VAEs) and generative adversarial networks (GANs), which lack of samples' variety and details.
Differently, a commonly used solution to generate novel datasets is based on synthetic rendering such as Unity\textsuperscript{\tiny\textregistered}~\cite{unity} and Unreal Engine\textsuperscript{\tiny\textregistered}~\cite{unrealengine} frameworks.
Unfortunately, those technologies often fail to provide realistic data, lacking of many realistic features such as accurate light reflections, camera artifacts, and noisy data.
As a result,  the data distribution of real samples will differ from synthesized ones, despite many works have been proposed to address the problem via domain adaptation and randomization approaches~\cite{ALKHALIFAH2022101, tobin2017domain, tremblay2018training}.

The lack of a large amount of ground truth data is particularly significant in the case of dense prediction applications, such as depth estimation, where RGB images and corresponding depth maps are required to perform the task.
This situation is likely related to the difficulties and highly time-consuming procedures needed to collect congruent RGB and depth data.
Such issues are not limited to calibration and alignment procedures between cameras and depth sensors but are also related to unfilled depth maps captured with LiDAR devices and the wide range of possible scenarios.
Even if many RGBD datasets have been proposed~\cite{lopes2022survey}, most of them include less than $50K$ real-world samples such as NYU Depth v2 (NYU)~\cite{nyu} and KITTI~\cite{kitti} datasets.
In contrast, millions of labeled samples are available for other computer vision tasks such as image classification (ImageNet~\cite{imagenet}) and object detection (COCO~\cite{coco}).
Consequently, the objective of this paper is to automatically generate realistic RGBD samples in order to increase the amount of training data while improving the deep learning model's performances, aiming to overcome the limits of data augmentation and synthetically created samples.
Our proposed solution, named Diffusion4D (D4D), is based on \textit{denoising diffusion probabilistic models} (DDPMs)~\cite{denoising_diff_mod, sohl2015deep}, a score-based generation techniques that have shown outstanding results in the creation of high-fidelity images~\cite{diff_mod_survey}.
Our strategy focuses on a custom 4-channels DDPM to capture the intrinsic information presents in real indoor and outdoor RGBD samples in order to generate realistic RGB images and corresponding depth maps while improving the data diversity between training samples.
D4D introduces customized architecture configurations which are based on 4-channels samples, fine-tuned loss functions, and diffusion schedules.
The designed models are used to drive the learning procedure of the DDPM to generate (unconditioned\footnote{The unconditioned generation techniques are identified by the absence of additional input data.}) heterogeneous variations of the original RGBD dataset.
Exploiting the characteristic of DDPMs based on the principle of non-equilibrium statistical physics, our aim is to extract key features of real RGBD samples during the forward (inference) process; subsequently, during the backward (generative) phase, the model generates realistic variations of original data obtained merging previously learned features.
Therefore, we do not target the production of highly photo-realistic images rather than coherent samples where RGB values and depth distances are correlated as in real-world; some examples are shown in Figure~\ref{fig:cover}. 
Furthermore, to demonstrate the effectiveness of generated RGBD samples, we apply D4D in a novel supervised training pipeline to tackle the monocular depth estimation (MDE)~\cite{ming2021deep} task, a dense prediction task consisting of estimating a per-pixel distance map given a single RGB image as input.

The main contributions of this work are summarized as follows:
\textbf{1)}~We design a customized 4-channels diffusion model to generate realistic RGBD samples.
\textbf{2)}~We incorporate D4D-generated data into a novel training pipeline to boost MDE models' performances.
\updated{\textbf{3)}~\upupupupdated{We demonstrate the effectiveness of the proposed training strategy to tackle the MDE task over four reference MDE models.
In particular, we focus on three convolution neural networks (CNN) and one hybrid vision transformer (hViT), which are respectively DenseDepth~\cite{densedepth}, FastDepth~\cite{fastdepth}, SPEED~\cite{speed}, and METER~\cite{10078346}.
We identify those architectures in order to provide a general overview of the adaptability of the proposed solution over various MDE architectures; precisely, in Section \ref{sec:experiments}, we will report the quantitative and qualitative estimation error reduction achieved with the employment of the D4D training pipeline over both indoor and outdoor scenarios.
Furthermore, we report some additional experiments on two efficient ViT architectures proposed in~\cite{effmeter}.
Subsequently}, we show the superior performances of generated samples in three settings: \textbf{3.1)}~When the training of MDE models is performed without the original dataset. 
\textbf{3.2)}~When compared against synthetic datasets, such as SceneNet RGB-D~\cite{scenenet} and SYNTHIA-SF~\cite{synthia} datasets.
\textbf{3.3)}~In generalization performances on the indoor DIML/CVL RGB-D~\cite{diml} test dataset in blind conditions.
\textbf{4)}~\upupupupdated{Finally, we created two new datasets, namely D4D-NYU and D4D-KITTI, each dataset refers to the original one (NYU, KITTI) and it is internally divided according to the generation resolution used. 
The datasets collect D4D-generated RGBD samples at a variety of resolutions, ranging from $64\times48$ pixels to $320\times240$ pixels. 
We hope that such datasets could be further exploited to improve the performances of MDE architectures and other depth-based tasks.
The project page and generated datasets are publicly available at the following link \href{https://github.com/lorenzopapa5/Diffusion4D}{\small{\texttt{https://github.com/lorenzopapa5/Diffusion4D}}}.}}

\modified{
This paper is organized as follows: Section~\ref{sec:related_works} reviews some previous works related to the topics of interest. Section~\ref{sec:methodology} describes the proposed D4D method and the overall training pipeline in detail.
Experiments and hyper-parameters are discussed in Section~\ref{sec:experimental_setup}, while Section~\ref{sec:experiments}  reports the qualitative and quantitative improvements achieved by the chosen MDE model with the use of D4D generated samples. Some final considerations and future applications are provided in Section~\ref{sec:conclusions}.
}

\section{Related Work}
\label{sec:related_works}
The task of producing new samples from an existing data collection is known as generation.
There are two basic generation methodologies: unconditioned, in which the samples are generated from noise (i.e., Gaussian noise), and conditioned, in which the samples are generated in response to  a given input, e.g., text prompts and images.
In AI-based approaches, this task is usually tackled through VAEs, GANs, and the recent DDPMs, deep learning techniques commonly based on convolutional and transformer operations. 
Many aspects in developing models for generating realistic images have been studied and improved during these years, such as conditioning the output with ad-hoc input variables as well as speeding up the process by working on the efficiency and inference frequency.
Zhu \emph{et al.}~\cite{zhu2019dm} (2019) propose DM-GAN, a text-conditioned architecture able to improve the quality of generated samples based on
information prompts.
Karras \emph{et al.}~\cite{karras2020training} (2020) focus on an augmentation solution for training a GAN model under limited data constraints.
Cai \emph{et al.}~\cite{CAI2020102697} (2020) propose a deep convolutional GAN solution to generate synthetic data to tackle the imbalanced problem of training datasets for crash prediction scenarios. 
Zhao \emph{et al.}~\cite{zhao2021improved} (2021) integrate and optimize the computational complexity of transformer architectures into a GAN-based approach in order to produce high-resolution images.

Furthermore, generative models have also been widely applied to handle the image translation task, in which an input image from one domain is translated (mapped) to another one while preserving the content of the given image.
An example is provided by Zhu \emph{et al.}~\cite{cyclegan} (2017) with CycleGAN, where the authors mainly focus on a cycle consistency loss to enhance the overall generation performances. 
Russo \emph{et al.}~\cite{russo2018source} (2018), inspired by \cite{cyclegan}, introduce a class consistency loss for cross-domain classification tasks.
Moreover, Tang~\emph{et al.}~\cite{tang2021attentiongan} (2021) propose to guide the translation process through an attention mechanism in order to achieve high-fidelity images, whereas Torbunov \emph{et al.}~\cite{torbunov2023uvcgan} (2023) improve CycleGAN performances by incorporating transformers layers as the generator. 
Similarly to previous related works and closer to our application scenario, Du \emph{et al.}~\cite{du2019translate} (2019) present a specific domain shift model to extract depth maps from RGB images. 
This work has been motivated by the limited amount of labeled data provided in existing RGBD datasets,

Recently, DDPMs~\cite{sohl2015deep}, a powerful new family of deep generative models have been proposed. 
Such architectures are based on two Markov chains: a forward chain that perturbs input data to noise and a reverse chain that translates noise to data. 
Ho \emph{et al.}~\cite{denoising_diff_mod} (2020) demonstrate DDPM capabilities in computer vision applications for the generation of high-quality images. 
Moreover, Dhariwal \emph{et al.}~\cite{NEURIPS2021_49ad23d1} (2021) shows that such models are able to achieve superior performances than GANs to handle image synthesis.
However, those architectures require substantial computational resources to be trained; consequently, Rombach \emph{et al.}~\cite{rombach2022high} (2022) propose a latent diffusion model that can be trained on limited computational resources proposing to integrate the Markovian structure into the latent space of a pretrained autoencoder network. 
Contrarily, Peebles \emph{et al.}~\cite{peebles2022scalable} (2022) replace the commonly-used U-Net \cite{ronneberger2015u} with transformer modules improving the generation capabilities while increasing the computational complexity. 

In contrast to such AI-based approaches, another popular solution for the generation of (potentially unlimited) samples is based on the extraction of frames and associated ground truth data from virtual environments, i.e., generated via graphical engines such as Unity\textsuperscript{\tiny\textregistered}, Unreal Engine\textsuperscript{\tiny\textregistered} and the most recent NVIDIA Isaac Sim\textsuperscript{\tiny\texttrademark} (Replicator)~\cite{nvidia_isaac}.
Those technologies often fail to provide realistic data, lacking artifact information commonly present in real-world images, resulting in poor performance at the inference step.
Synthetic datasets, generated with graphic engines, have been widely employed in the MDE task. 
Zou \emph{et al.}~\cite{zou2018df} (2018) use the synthetic SYNTHIA datasets as a pre-training strategy to improve depth estimation performances on autonomous driving scenarios, while Chen \emph{et al.}~\cite{chen2019learning} (2019) employ the synthetic SceneNet dataset to increase the number of training samples and the model's generalization performances.
Contrarily, Xian \emph{et al.}~\cite{xian2020structure} (2020) propose to estimate pseudo-depth data trained on relative depth datasets to improve the model's generalization in real-world scenarios.
The work also underlies the presence of a domain gap between synthetic and real data, as well as the need for domain adaptation techniques to efficiently use synthesized samples.

Consequently, based on similar motivation of~\cite{du2019translate, xian2020structure}, in this paper, we integrate in a novel training pipeline a custom 4-channels DDPM in order to generate realistic RGBD samples for both indoor and outdoor contexts and improve the estimation performances of MDE approaches while overcoming the limitations introduced by graphical engines.
To the best of our knowledge, no previous works propose a similar solution to improve a dense prediction task; a detailed description of the proposed training pipeline is following reported.


\begin{figure*}
    \centering
    \includegraphics[width=\linewidth]{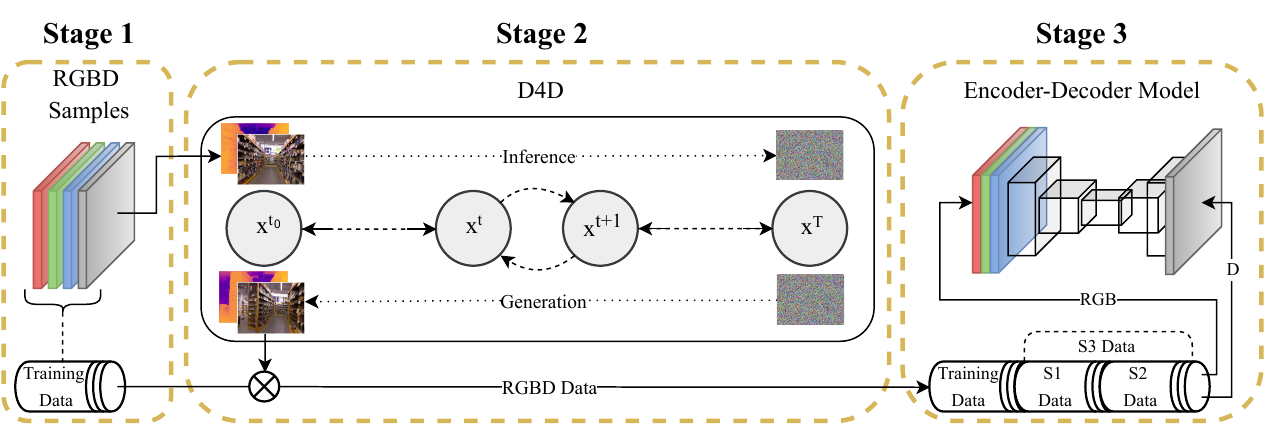}   
    \caption{\upupdated{Graphical representation of the introduced training pipeline. \textbf{Stage 1} shows the pre-processing operations applied on 4-channels samples extracted from the original training dataset. \textbf{Stage 2} emphasizes the training and unconditioned generation processes of D4D model. \textbf{Stage 3} depicts the training procedure of a generic encoder-decoder MDE network by highlighting how the RGBD training samples are composed.}}
    \label{fig:pipeline}
\end{figure*}

\section{Methodology}
\label{sec:methodology}
\updated{
This section describes the proposed pipeline for generating RGBD samples with the D4D model.
As mentioned in Section \ref{sec:introduction}, one of the primary bottlenecks in the MDE task, a computer vision application where a dense depth map is predicted from a single RGB image, is the lack of a large amount of training data.
Therefore, the proposed training pipeline aims to improve the estimation performances of well-known MDE architectures by generating RGBD samples learned from real-world 4-channels (images) data distribution.
We report a graphical representation in Figure~\ref{fig:pipeline}; as can be seen, the pipeline is divided into three stages described below. 

\vspace{1em}
\noindent
\textbf{Stage 1:} The first phase is characterized by widely employed preprocessing techniques.
More in detail, we select as training datasets the NYU for the indoor scenarios and the KITTI for the outdoor ones, both of which are composed of real-world RGBD samples. 
Furthermore, the pixel values of the training samples are normalized into the $[0,1]$ range and rescaled to the working model resolution.
Consequently, the image's height and width are scaled \upupdated{(resized with a bilinear interpolation process)} to the working resolutions of the compared architectures used in Stage 3, such as $640\times480$ (DenseDepth), $224\times224$ (FastDepth), and $256\times192$ (SPEED and METER). 
This choice will influence (in Stage 2) the generation resolution of D4D model at inference time.

\vspace{1em}
\noindent
\textbf{Stage 2:} The second phase is devoted to generating realistic samples; precisely, we leverage our custom DDPM to produce 4-channels samples based on the original training data.}
Before introducing our generation strategy, let us briefly review some basic concepts necessary to better understand DDPMs, highlighting the motivations that led us to develop the proposed solutions.
DDPMs, inspired by non-equilibrium statistical physics, exploit the reduction of the input data distribution into a well-known one, in our case, the Gaussian distribution. 
This process, known as forward diffusion (inference), is then reversed (generation) to restore input data distribution. 
This procedure is commonly defined in literature as \textit{highly flexible} and \textit{tractable} since the model can potentially represent unlimited data distributions. 
According to this behavior, the straightforward baseline idea of this paper is to use a DDPM to learn the distribution of RGBD data from real-world benchmark datasets during the forward phase.
As a result, during the generation phase, D4D could produce multiple realistic 4-channels variations of original ground-truth data by combining previously extracted features.

\updated{Therefore, we introduce some basic knowledge about diffusion model methodologies by focusing on the main parameters that would impact D4D generation performance. 
More in detail, diffusion models are characterized by forward and reverse procedures.
The training process of our diffusion model is principally driven by the cost function $L(\cdot, \cdot)$ and the diffusion rate $\beta$.
The first function, usually a $L1(\cdot, \cdot)$ (mean-absolute) or $L2(\cdot, \cdot)$ (mean-squared) loss, is computed between the input data distribution $q(x^{t_0})$ and the generated one $p(x^{t_0})$ to fit the DDPM data distribution $\pi(y)$, which usually represents a Gaussian distribution.
At the forward phase, the diffusion rate, as defined in~\cite{sohl2015deep}, drives the Markov diffusion kernel $t_{\pi}(y|y';\beta_t)$ with $t = [t_0; T]$ steps, to make the distribution $\pi(y)$ analytically tractable, while the reverse phase is trained to describe the same trajectory, but in a reverse way; we report the two procedures in the following equations.
\upupupupdated{
\begin{equation}
    forward \rightarrow q(x^{t}) = q(x^{t_0}) \Pi_{t_0}^T t_{\pi}(x|x';\beta_t)
\end{equation}
\begin{equation}
    reverse \rightarrow p(x^{t}) = \pi(x^{T}) \Pi_{t_0}^T t_{\pi}(x'|x)     
\end{equation}}
}
Moreover, the configuration of the diffusion rate is fundamental for its final performances; in~\cite{denoising_diff_mod, sohl2015deep} authors set a \textit{linear} $\beta$ variance ranging from $\beta_1 = 10^{-4}$ to $\beta_T=0.02$ with $T=1000$. In contrast, in \cite{nichol2021improved}, authors propose to improve diffusion models with a reparametrization of the generation process variance, i.e., replacing the linear schedule with a squared \textit{cosine} to prevent abrupt changes of noise levels. 
This choice leads to a slower forward process with $T=4000$ steps while increasing reconstructed image details.

\upupupdated{Based on the just introduced description on diffusion model methodologies and influenced by the loss function formulation commonly employed in the MDE task~\cite{densedepth, acc_obj_loss}, where the learning process usually relies on multiple loss functions focused on contours, fine details, and images as a whole, we design D4D with a similar behavior.
Precisely, the proposed strategy would combine two configurations of loss functions and beta scheduler setups in order to ensure diversity and consistency in the generated RGBD samples.
\upupupupdated{The combination of diversity and consistency of the generated samples, which are combined into the training set, act as a powerful and realistic data augmentation schema, which is able to increase the generalization capabilities of our network, resulting in a lower testing error as shown in the Results section.
More in detail, we propose a merging strategy based on two complementary configurations, namely S1 and S2, that are able to generate realistic samples with various data distributions in order to enhance the overall depth estimation performances of well-known MDE models.}}
In the first configuration (S1), the model focuses on creating realistic images mainly composed of constant or gradually increasing depth distances. 
\upupupdated{As a result, we develop S1 with a slow convergence behavior, i.e., characterized by an $L1$ loss function to mitigate the error during the training process, and a linear diffusion rate ($\beta$)~\cite{denoising_diff_mod, sohl2015deep} leading the model to a faster forward process with the constant addition of noisy data. 
Moreover, by defining with $\mathcal{P}$ the set of pixels, for any pixel $p\in \mathcal{P}$, the S1 configuration can be formalized as \upupupupdated{reported in Equation~\ref{eq:s1}.
\begin{equation}
    \label{eq:s1}
    S1: L1 = \frac{1}{|\mathcal{P}|} \sum_{p \in \mathcal{P}} || x_p - y_p ||_1,  \hspace{0.5em} \beta = linear
\end{equation}}}
In contrast, in the second configuration (S2), we look for generated images that are rich in detail with stronger distance variations.
\upupupdated{Consequently, we implement S2 with a slower forward process better focusing on details and objects in the images, i.e., a cosinusoidal diffusion scheme ($\beta$)~\cite{nichol2021improved} combined with a $L2$ loss function to achieve a fast convergence of the learning system.
Moreover, by defining with $\mathcal{P}$ the set of pixels, for any pixel $p\in \mathcal{P}$, the S2 configuration can be formalized as \upupupupdated{reported in Equation~\ref{eq:s2}.
\begin{equation}
    \label{eq:s2}
    S2: L2 = \frac{1}{|\mathcal{P}|} \sum_{p \in \mathcal{P}} || x_p - y_p ||_2^2, \hspace{0.5em} \beta = cosine
\end{equation}}}

\upupupdated{\upupupupdated{Finally, the proposed configuration (S3) is composed by merging the generated RGBD samples from S1 and S2.
We opted to set the number of steps $T$ equal to $1000$ as a trade-off between training time and image photorealism.
Under these settings, S3 effectively encompasses a wide range of possible RGB and depth data distributions while balancing the convergence speed and the diffusion rate of the 4-channels DDPM.
Moreover, by defining with $s1$ and $s2$ the set of generated RGBD data, respectively, from S1 and S2 configurations, the proposed strategy can be summarized as follows:}
\upupupupdated{
\begin{equation}
    S3 = (s1 \cup s2) \hspace{0.3em} where \hspace{0.3em} \begin{cases}
    S1:\{ loss: L1, & \beta: linear \}\\
    S2:\{ loss: L2, & \beta: cosine \}
    \end{cases} 
\end{equation}}}

We conclude this stage by merging the generated RGBD samples with the original training data in order to create a unique augmented training set.
\modified{
Furthermore, because DDPM has a significant computing cost during the training and generation stages, we perform all of the operations described in this step offline.
}

\vspace{1em}
\noindent
\updated{\textbf{Stage 3:} Following the proposed training pipeline, in the last phase, we employ the novel augmented training set to tackle the MDE task.
Precisely, we employ the RGB images and respective depth maps to train commonly used encoder-decoder architectures, which are represented as transparent blocks in Figure \ref{fig:pipeline}; in particular, we focus on DenseDepth, FastDepth, SPEED, and METER, which are typically deep and shallow architectures commonly used in the MDE task.
We chose these models due to their different working resolutions, architectural components, and estimation capabilities in order to demonstrate the effectiveness of D4D-generated samples at different scales and performances.}
This final phase is fundamental for demonstrating the efficacy of the proposed training pipeline and for quantitatively measuring the attained improvement.

\modified{
\section{Experimental Setup}
\label{sec:experimental_setup}
In this section, we describe hyperparameter setups of trained architectures and  evaluation metrics used to compare their performances.
} 
The proposed method is implemented on the PyTorch framework~\cite{paszke2019}.
To generate new samples with the D4D procedure, we employ two benchmark MDE datasets, i.e., NYU Depth v2 and KITTI, following the Eigen \emph{et al.}~\cite{eigen2014depth} (2014) split strategy.
NYU and KITTI are respectively composed of around $(50K, 23K)$ training and $(654, 652)$ test samples at a resolution of $(640\times480, 1242\times375)$ and a maximum depth range of ($10$, $80$) meters.
Furthermore, to compare the performances achieved by generated samples with respect to synthetic ones (Figure \ref{fig:pipeline}, Stage 3), we use the SceneNet dataset for the indoor scenario and the SYNTHIA-SF for the outdoor one.
We use a subset of $300K$ samples for the first dataset and the entire training set for the second one, composed of $3K$ samples.
Finally, we use the $503$ samples of the DIML test dataset to show the generalization performances on an unseen set of data.
Moreover, following the training pipeline outlined in the previous section, we describe the hyperparameters and evaluation metrics used in this paper.

In Stage 2, we train each configuration (S1 and S2) at different image resolutions ranging from $64\times48$ pixels to $320\times240$ pixels on NYU and KITTI datasets. 
The DDPM layers are initialized as described in~\cite{denoising_diff_mod, qiao2019micro}.
We train D4D for $150$ epochs with a batch size ranging from $256$ to $16$ depending on the image resolution on an NVIDIA A100 SXM4.
We use Adam as optimizer with decoupled weight decay~\cite{loshchilov2017decoupled} of $1\times10^{-2}$, a learning rate equal to $1\times10^{-4}$ and a decay of $1\times10^{-1}$ after $100$ and $125$ epochs.
Following common practice we set remaining hyperparameters as $\beta_1 = 0.9$, $\beta_2 = 0.999$ and $\epsilon=1\times10^{-8}$. 

\updated{In Stage 3, we train all the compared MDE models (DenseDepth, FastDepth, SPEED, and METER) with the following hyperparameter setting: we use Adam optimizer configuration as before with a learning rate equal to $1\times10^{-3}$ and a decay of $1\times10^{-1}$ every $20$ epochs for a total of $80$ epochs on an NVIDIA RTX 3090.  
Furthermore, we initialized the convolutional kernels as suggested in respective papers \cite{densedepth, fastdepth, speed, 10078346} and trained/tested the MDE architectures with original input-output model resolutions, i.e., $(640\times480, 320\times240)$, $(224\times224, 224\times224)$, $(256\times192, 64\times48)$ and $(256\times192, 64\times48$ or $640\times192, 160\times48)$\footnote{Differently to the other compared CNN architecture, METER has different image resolutions between the indoor and outdoor scenarios (same height but different width).} respectively for DenseDepth, FastDepth, SPEED, and METER.}
The training procedure is further enriched using the strategy proposed in~\cite{densedepth} with the addition of the random crop. 
Finally, we evaluate the trained models following the evaluation metrics introduced in~\cite{eigen2014depth}: root mean squared error (RMSE, in meters [m]), mean absolute error (MAE, in meters [m]), absolute relative error (Abs$_{Rel}$), and accuracy values such as $\delta_1$,  $\delta_2$ and $\delta_3$.
\modified{
Moreover, for any pixel $p\in \mathcal{P}$, we define its ground truth depth map as $y_p$ while $\hat{y}_p$ is the predicted one.
Those evaluation metrics are formally defined in the following \upupupupdated{equations.
\begin{equation}
    RMSE = \sqrt{\frac{1}{|\mathcal{P}|} \sum_{p \in \mathcal{P}} || y_p - \hat{y}_p ||^2}
\end{equation}
\begin{equation}
    MAE = \frac{1}{|\mathcal{P}|} \sum_{p \in \mathcal{P}} | y_p - \hat{y}_p |
\end{equation}
\begin{equation}
    Abs_{Rel} = \frac{1}{|\mathcal{P}|} \sum_{p \in \mathcal{P}} \frac{|y_p - \hat{y}_p |}{y_p}
\end{equation}
}

\noindent
For estimating the accuracy values $\delta_{z\in \mathbf{N}}$ with $z\in[1,3]$, a threshold ($thr$) is commonly set to $1.25^z$ while the set of pixel $\mathcal{P}_{z}^{*}$ is defined as follows:
\upupupupdated{
\begin{equation}
    \mathcal{P}^*_z = \biggl\{ p\in \mathcal{P}\ s.t. \max \left(\frac{y_p}{\hat{y}_p}, \frac{\hat{y}_p}{y_p}\right) < thr^z \biggr\}
\end{equation}
}

\noindent
Finally, the accuracy values can be expressed as \upupupupdated{reported in Equation~\ref{eq:d1}.
\begin{equation}
    \label{eq:d1}
    \delta_{z\in \mathbf{N}, z\in[1,3]} = \frac{|\mathcal{P}_z^*|}{|\mathcal{P}|}
\end{equation}}}

\begin{table*}[t]
    \footnotesize
    \centering
    \caption{Quantitative evaluation of different MDE architectures and configurations. The original samples are taken from NYU dataset \modified{(third column, NYU = 50K)}, the \textit{Synthetic} samples are from SceneNet, while the generated samples (\textit{Add}) are from D4D-NYU. The proposed S3 configuration is in bold, while the optimal strategy for each compared model is highlighted in gray.}
    \begin{center}
        \begin{tabular}{ c | c | c c c | c c c | c c c }
        \hline
        Model & Configuration & NYU [K] & \textit{Add} [K] & \textit{Res} [pix] & \cellcolor{orange!20}RMSE$\downarrow$ [m] & \cellcolor{orange!20}MAE$\downarrow$ [m] & \cellcolor{orange!20}Abs$_{Rel}$$\downarrow$ & \cellcolor{blue!15}$\delta_1$$\uparrow$ & \cellcolor{blue!15}$\delta_2$$\uparrow$ & \cellcolor{blue!15}$\delta_3$$\uparrow$ \\
        \hline\hline
        \multirow{9}*{DenseDepth} & S0 & 50 & 0 & - & 0.5021 & 0.3663 & 0.1445 & 0.8087 & 0.9507 & 0.9846 \\\cline{2-11}
         & \textit{Synthetic} & 50 & 50 & $320\times240$ & 0.4882 & 0.3438 & 0.1367 & 0.8199 & 0.9583 & 0.9888 \\
         & \textit{Synthetic} & 50 & 150 & $320\times240$ & 0.4713 & 0.3487 & 0.1358 & 0.8251 & 0.9634 & 0.9910 \\\cline{2-11}
         & S1 & 50 & 50 & $320\times240$ & 0.4575 & 0.3352 & 0.1294 & 0.8379 & 0.9640 & 0.9899 \\
         & S2 & 50 & 50 & $320\times240$ & 0.4598 & 0.3354 & 0.1273 & 0.8390 & 0.9667 & 0.9921 \\
         & \cellcolor{gray!30}\textbf{S3} & \cellcolor{gray!30}50 & \cellcolor{gray!30}50 & \cellcolor{gray!30}$320\times240$ & \cellcolor{gray!30}0.4568 & \cellcolor{gray!30}0.3368 & \cellcolor{gray!30}0.1327 & \cellcolor{gray!30}0.8340 & \cellcolor{gray!30}0.9659 & \cellcolor{gray!30}0.9912 \\
         & \cellcolor{gray!30}\textbf{S3} & \cellcolor{gray!30}50 & \cellcolor{gray!30}100 & \cellcolor{gray!30}$320\times240$ & \cellcolor{gray!30}0.4480 & \cellcolor{gray!30}0.3262 & \cellcolor{gray!30}0.1236 & \cellcolor{gray!30}0.8499 & \cellcolor{gray!30}0.9693 & \cellcolor{gray!30}0.9923 \\
         & \textbf{S3} & 50 & 50 & $256\times192$ & 0.4788 & 0.3513 & 0.1340 & 0.8241 & 0.9614 & 0.9912 \\
         & \textbf{S3} & 50 & 100 & $256\times192$ & 0.4578 & 0.3364 & 0.1286 & 0.8376 & 0.9672 & 0.9917 \\
        \hline\hline
        \multirow{8}*{FastDepth} & S0 & 50 & 0 & - & 0.5714 & 0.4317 & 0.1751 & 0.7535 & 0.9374 & 0.9820 \\\cline{2-11}
         & \textit{Synthetic} & 50 & 100 & $320\times240$ & 0.5468 & 0.4122 & 0.1617 & 0.7747 & 0.9450 & 0.9858 \\
         & \textit{Synthetic} & 50 & 300 & $320\times240$ & 0.5198 & 0.3883 & 0.1519 & 0.7948 & 0.9533 & 0.9870 \\\cline{2-11}
         & S1 & 50 & 100 & $256\times192$ & 0.5029 & 0.3741 & 0.1455 & 0.8058 & 0.9586 & 0.9892 \\
         & S2 & 50 & 100 & $256\times192$ & 0.5313 & 0.3995 & 0.1600 & 0.7775 & 0.9454 & 0.9869 \\
         & \cellcolor{gray!30}\textbf{S3} & \cellcolor{gray!30}50 & \cellcolor{gray!30}100 & \cellcolor{gray!30}$256\times192$ & \cellcolor{gray!30}0.4980 & \cellcolor{gray!30}0.3678 & \cellcolor{gray!30}0.1414 & \cellcolor{gray!30}0.8119 & \cellcolor{gray!30}0.9603 & \cellcolor{gray!30}0.9901 \\
         & \textbf{S3} & 50 & 50 & $320\times240$ & 0.5132 & 0.3810 & 0.1467 & 0.8014 & 0.9553 & 0.9886 \\
         & \textbf{S3} & 50 & 100 & $320\times240$ & 0.5103 & 0.3802 & 0.1492 & 0.7903 & 0.9507 & 0.9865 \\
        \hline\hline
        \multirow{8}*{SPEED} & S0 & 50 & 0 & - & 0.5638 & 0.4275 & 0.1676 & 0.7601 & 0.9357 & 0.9836 \\\cline{2-11}
         & \textit{Synthetic} & 50 & 100 & $320\times240$ & 0.5606 & 0.4247 & 0.1657 & 0.7605 & 0.9404 & 0.9857 \\
         & \textit{Synthetic} & 50 & 300 & $320\times240$ & 0.5542 & 0.4217 & 0.1633 & 0.7696 & 0.9496 & 0.9864 \\\cline{2-11}
         & S1 & 50 & 100 & $256\times192$ & 0.5170 & 0.3877 & 0.1482 & 0.7948 & 0.9549 & 0.9897 \\
         & S2 & 50 & 100 & $256\times192$ & 0.5216 & 0.3943 & 0.1486 & 0.7905 & 0.9565 & 0.9912 \\
         & \cellcolor{gray!30}\textbf{S3} & \cellcolor{gray!30}50 & \cellcolor{gray!30}100 & \cellcolor{gray!30}$256\times192$ & \cellcolor{gray!30}0.4982 & \cellcolor{gray!30}0.3712 & \cellcolor{gray!30}0.1430 & \cellcolor{gray!30}0.8054 & \cellcolor{gray!30}0.9610 & \cellcolor{gray!30}0.9911 \\
         & \textbf{S3} & 50 & 50 & $320\times240$ & 0.5132 & 0.3870 & 0.1494 & 0.7973 & 0.9559 & 0.9885 \\
         & \textbf{S3} & 50 & 100 & $320\times240$ & 0.5001 & 0.3767 & 0.1441 & 0.8090 & 0.9587 & 0.9903 \\
         \hline\hline
         \updated{\multirow{8}*{METER}} & \updated{S0} & \updated{50} & \updated{0} & \updated{-}  & \updated{0.5112} &  \updated{0.3854} & \updated{0.1439} & \updated{0.8138} & \updated{0.9577} & \updated{0.9876} \\\cline{2-11}
         & \updated{\textit{Synthetic}} & \updated{50} &  \updated{100} & \updated{$320\times240$} & \updated{0.4893} & \updated{0.3675} & \updated{0.1446} & \updated{0.8130} & \updated{0.9592} & \updated{0.9890} \\
         & \updated{\textit{Synthetic}} & \updated{50} &  \updated{300} & \updated{$320\times240$} & \updated{0.4957} & \updated{0.3709} & \updated{0.1446} & \updated{0.8150} & \updated{0.9574} & \updated{0.9882} \\\cline{2-11}
         & \updated{S1} & \updated{50} &  \updated{100} & \updated{$256\times192$} & \updated{0.4649} & \updated{0.3471} & \updated{0.1353} & \updated{0.8320} & \updated{0.9685} & \updated{0.9915} \\
         & \updated{S2} & \updated{50} &  \updated{100} & \updated{$256\times192$} & \updated{0.4760} & \updated{0.3584} & \updated{0.1388} & \updated{0.8202} & \updated{0.9660} & \updated{0.9923} \\
         & \cellcolor{gray!30}\textbf{S3} & \cellcolor{gray!30}50 &  \cellcolor{gray!30}100 & \cellcolor{gray!30}$256\times192$ & \cellcolor{gray!30}0.4574 & \cellcolor{gray!30}0.3390 & \cellcolor{gray!30}0.1290 & \cellcolor{gray!30}0.8357 & \cellcolor{gray!30}0.9667 & \cellcolor{gray!30}0.9924 \\
         & \updated{\textbf{S3}} & \updated{50} &  \updated{50} & \updated{$320\times240$} & \updated{0.4669} & \updated{0.3495} & \updated{0.1334} & \updated{0.8303} & \updated{0.9673} & \updated{0.9923} \\
         & \updated{\textbf{S3}} & \updated{50} &  \updated{100} & \updated{$320\times240$} & \updated{0.4615} & \updated{0.3447} & \updated{0.1320} & \updated{0.8350} & \updated{0.9695} & \updated{0.9928} \\
         \hline
        \end{tabular}
    \end{center}
    \label{tab:nyu_test_siNYU}
\end{table*}
\begin{table*}[!h]
    \footnotesize
    \centering
    \caption{Quantitative evaluation of different MDE architectures and configurations. The \textit{Synthetic} samples are from SceneNet while the generated samples (\textit{Add}) are from D4D-NYU \modified{while no NYU (original) samples are used \modified{(third column, NYU = 0K)}}. The proposed S3 configuration is in bold, while the optimal strategy for each compared model is highlighted in gray.}
    \begin{center}
        \begin{tabular}{ c | c | c c c | c c c | c c c }
        \hline
        Model & Configuration & NYU [K] & \textit{Add} [K] & \textit{Res} [pix] & \cellcolor{orange!20}RMSE$\downarrow$ [m] & \cellcolor{orange!20}MAE$\downarrow$ [m] & \cellcolor{orange!20}Abs$_{Rel}$$\downarrow$ & \cellcolor{blue!15}$\delta_1$$\uparrow$ & \cellcolor{blue!15}$\delta_2$$\uparrow$ & \cellcolor{blue!15}$\delta_3$$\uparrow$ \\
        \hline\hline
        \multirow{9}*{DenseDepth} & S0 & 0 & 0 & - & - & - & - & - & - & - \\\cline{2-11}
         & \textit{Synthetic} & 0 & 50 & $320\times240$ & 1.1034 & 0.8648 & 0.4298 & 0.4123 & 0.6886 & 0.8465 \\
         & \textit{Synthetic} & 0 & 150 & $320\times240$ & 1.0383 & 0.8292 & 0.4019 & 0.4197 & 0.7228 & 0.8825 \\\cline{2-11}
         & S1 & 0 & 50  & $320\times240$ & 0.5559 & 0.4250 & 0.1736 & 0.7549 & 0.9373 & 0.9821 \\
         & S2 & 0 & 50 & $320\times240$ & 0.6087 & 0.4767 & 0.1931 & 0.6619 & 0.9297 & 0.9773 \\
         & \cellcolor{gray!30}\textbf{S3} & \cellcolor{gray!30}0 & \cellcolor{gray!30}50 & \cellcolor{gray!30}$320\times240$ & \cellcolor{gray!30}0.5306 & \cellcolor{gray!30}0.4030 & \cellcolor{gray!30}0.1580 & \cellcolor{gray!30}0.7755 & \cellcolor{gray!30}0.9489 & \cellcolor{gray!30}0.9873\\
         & \cellcolor{gray!30}\textbf{S3} & \cellcolor{gray!30}0 & \cellcolor{gray!30}100 & \cellcolor{gray!30}$320\times240$ & \cellcolor{gray!30}0.5301 & \cellcolor{gray!30}0.4003 & \cellcolor{gray!30}0.1578 & \cellcolor{gray!30}0.7754 & \cellcolor{gray!30}0.9490 & \cellcolor{gray!30}0.9873\\
         & \textbf{S3} & 0 & 50 & $256\times192$  & 0.5473 & 0.4163 & 0.1654 & 0.7654 & 0.9446 & 0.9866 \\
         & \textbf{S3} & 0 & 100 & $256\times192$ & 0.5398 & 0.4096 & 0.1597 & 0.7720 & 0.9469 & 0.9877 \\
        \hline\hline
        \multirow{8}*{FastDepth} & S0 & 0 & 0 & - & - & - & - & - & - & - \\\cline{2-11}
         & \textit{Synthetic} & 0 & 100 & $320\times240$ & 1.1169 & 0.9779 & 0.4538 & 0.3866 & 0.6903 & 0.8621 \\
         & \textit{Synthetic} & 0 & 300 & $320\times240$ & 1.0852 & 0.9051 & 0.4167 & 0.
         4247 & 0.7275 & 0.8817 \\\cline{2-11}
         & S1 & 0 & 100 & $256\times192$ & 0.5709 & 0.4319 & 0.1768 & 0.7543 & 0.9412 & 0.9839 \\
         & S2 & 0 & 100 & $256\times192$ & 0.5952 & 0.4569 & 0.1845 & 0.7047 & 0.9292 & 0.9842 \\
         & \cellcolor{gray!30}\textbf{S3} & \cellcolor{gray!30}0 & \cellcolor{gray!30}100 & \cellcolor{gray!30}$256\times192$ & \cellcolor{gray!30}0.5502 & \cellcolor{gray!30}0.4165 & \cellcolor{gray!30}0.1730 & \cellcolor{gray!30}0.7649 & \cellcolor{gray!30}0.9464 & \cellcolor{gray!30}0.9877 \\
         & \textbf{S3} & 0 & 50 & $320\times240$ & 0.5735 & 0.4397 & 0.1756 & 0.7468 & 0.9389 & 0.9844 \\
         & \textbf{S3} & 0 & 100 & $320\times240$ & 0.5651 & 0.4343 & 0.1721 & 0.7473 & 0.9394 & 0.9854 \\
        \hline\hline
        \multirow{8}*{SPEED} & S0 & 0 & 0 & - & - & - & - & - & - & - \\\cline{2-11}
         & \textit{Synthetic} & 0 & 100 & $320\times240$ & 1.2278 & 1.0606 & 0.5424 & 0.3159 & 0.6279 & 0.8290 \\
         & \textit{Synthetic} & 0 & 300 & $320\times240$ & 1.1635 & 0.9827 & 0.4732 & 0.3923 & 0.6850 & 0.8532 \\\cline{2-11}
         & S1 & 0 & 100 & $256\times192$ & 0.5833 & 0.4430 & 0.1687 & 0.7493 & 0.9385 & 0.9857 \\
         & S2 & 0 & 100 & $256\times192$ & 0.6003 & 0.4646 & 0.1779 & 0.6875 & 0.9224 & 0.9825 \\
         & \cellcolor{gray!30}\textbf{S3} & \cellcolor{gray!30}0 & \cellcolor{gray!30}100 & \cellcolor{gray!30}$256\times192$ & \cellcolor{gray!30}0.5590 & \cellcolor{gray!30}0.4260 & \cellcolor{gray!30}0.1622 & \cellcolor{gray!30}0.7665 & \cellcolor{gray!30}0.9438 & \cellcolor{gray!30}0.9874 \\
         & \textbf{S3} & 0 & 50 & $320\times240$ & 0.5803 & 0.4482 & 0.1735 & 0.7456 & 0.9352 & 0.9852 \\
         & \textbf{S3} & 0 & 100 & $320\times240$ & 0.5694 & 0.4379 & 0.1674 & 0.7439 & 0.9423 & 0.9862 \\
         \hline\hline
         \updated{\multirow{8}*{METER}} & \updated{S0} & \updated{0} &  \updated{0} & \updated{-} & \updated{-} & \updated{-} & \updated{-} & \updated{-} & \updated{-} & \updated{-} \\\cline{2-11}
         & \updated{\textit{Synthetic}} & \updated{0} &  \updated{100} & \updated{$320\times240$} & \updated{1.2242} & \updated{1.0100} & \updated{0.4319} & \updated{0.3688} & \updated{0.6770} & \updated{0.8547} \\
         & \updated{\textit{Synthetic}} & \updated{0} &  \updated{300} & \updated{$320\times240$} & \updated{1.0480} & \updated{0.8556} & \updated{0.3837} & \updated{0.4468} & \updated{0.7403} & \updated{0.8909} \\\cline{2-11}
         & \updated{S1} & \updated{0} &  \updated{100} & \updated{$256\times192$} & \updated{0.5445} & \updated{0.4140} & \updated{0.1636} & \updated{0.7679} & \updated{0.9474} & \updated{0.9863}\\
         & \updated{S2} & \updated{0} &  \updated{100} & \updated{$256\times192$} & \updated{0.5905} & \updated{0.4574} & \updated{0.1837} & \updated{0.7180} & \updated{0.9322} & \updated{0.9851} \\
         & \cellcolor{gray!30}\textbf{S3} & \cellcolor{gray!30}0 & \cellcolor{gray!30}100 & \cellcolor{gray!30}$256\times192$ & \cellcolor{gray!30}0.5370 & \cellcolor{gray!30}0.4075 & \cellcolor{gray!30}0.1577 & \cellcolor{gray!30}0.7711 & \cellcolor{gray!30}0.9510 & \cellcolor{gray!30}0.9886 \\
         & \updated{\textbf{S3}} & \updated{0} &  \updated{50} & \updated{$320\times240$} & \updated{0.5778} & \updated{0.4465} & \updated{0.1709} & \updated{0.7729} & \updated{0.9366} & \updated{0.9862} \\
         & \updated{\textbf{S3}} & \updated{0} &  \updated{100} & \updated{$320\times240$} & \updated{0.5368} & \updated{0.4125} & \updated{0.1602} & \updated{0.7686} & \updated{0.9491} & \updated{9887} \\
         \hline
        \end{tabular}
    \end{center}
    \label{tab:nyu_test_noNYU}
\end{table*}

\begin{figure*}[t]
    \centering
    \includegraphics[width=\linewidth]{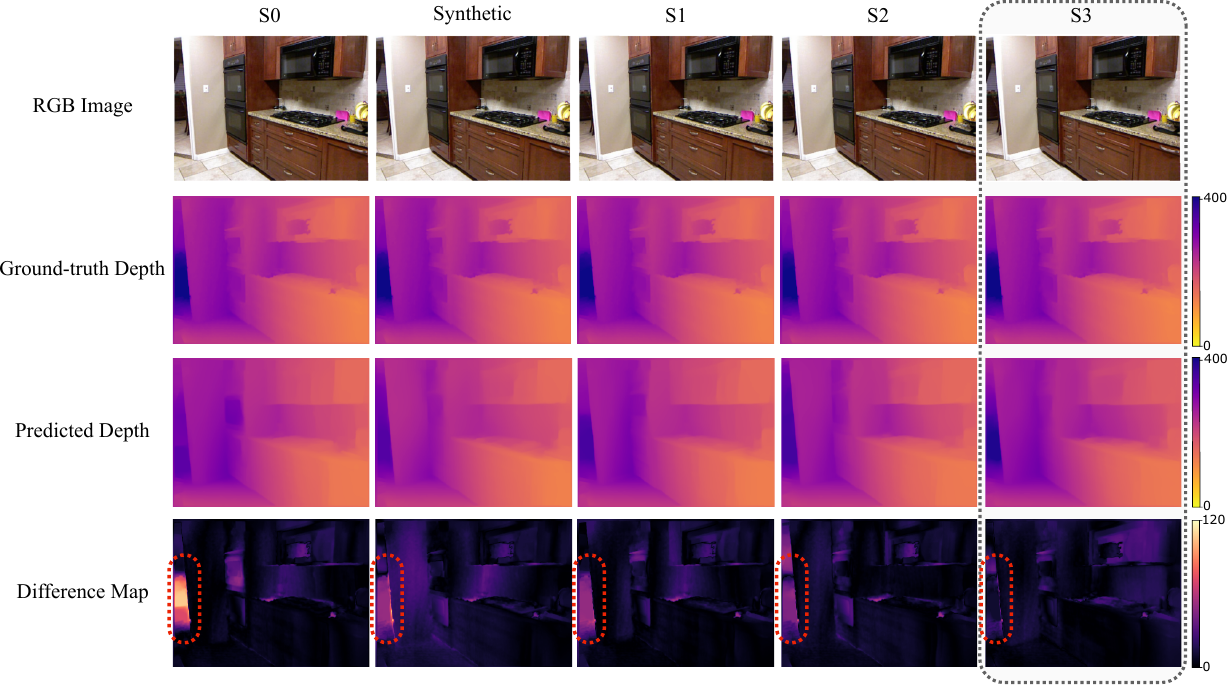} 
    \caption{\modified{\textbf{Indoor results.} Qualitative analysis of the estimated prediction obtained with DenseDepth method. The model has been tested on NYU (indoor) dataset. S0 is the baseline setup, i.e., when the MDE model is trained only on the NYU dataset. In \textit{Synthetic} setup, DenseDepth has been trained over NYU and a $50K$ subset from the SceneNet dataset. In S\textit{i} with $i=[1, 3]$, as described in Section 3,  DenseDepth has been trained over NYU and $50K$ samples taken from our proposed D4D-NYU datasets generated at a resolution of $320\times240$. The Difference Map is computed as a per pixel-difference between predicted ($\hat{y}$) and expected depth ($y$), while the reported colorbars are used to emphasize the depth/error range in centimeters ($cm$).}}
    \label{fig:densedepth_indoor}
\end{figure*}
\begin{table*}[!h]
    \footnotesize
    \centering
    \caption{Quantitative evaluation of different MDE architectures and configurations. The original samples are taken from KITTI dataset, the \textit{Synthetic} samples are from SYNTHIA-SF while the generated samples (\textit{Add}) are from D4D-KITTI. The proposed S3 configuration is in bold, while the optimal strategy for each compared model is highlighted in gray.}
    \begin{center}
        \begin{tabular}{ c | c | c c c | c c c | c c c }
        \hline
        Model & Configuration & KITTI [K] & \textit{Add} [K] & \textit{Res} [pix] & \cellcolor{orange!20}RMSE$\downarrow$ [m] & \cellcolor{orange!20}MAE$\downarrow$ [m] & \cellcolor{orange!20}Abs$_{Rel}$$\downarrow$ & \cellcolor{blue!15}$\delta_1$$\uparrow$ & \cellcolor{blue!15}$\delta_2$$\uparrow$ & \cellcolor{blue!15}$\delta_3$$\uparrow$ \\
        \hline\hline
        \multirow{6}*{DenseDepth} & S0 & 23 & 0 & - & 5.2099 & 3.1749 & 0.1417 & 0.7991 & 0.9475 & 0.9840 \\\cline{2-11}
         & \textit{Synthetic} & 23 & 3 & $1940\times1080$ & 5.2982 & 3.2499 & 0.1448 & 0.7871 & 0.9458 & 0.9856 \\\cline{2-11}
         & S1 & 23 & 50 & $320\times240$ & 5.1284 & 3.0221 & 0.1341 & 0.8057 & 0.9546 & 0.9882 \\
         & S2 & 23 & 50 & $320\times240$ & 5.1437 & 3.0539 & 0.1349 & 0.7989 & 0.9533 & 0.9869 \\
         & \cellcolor{gray!30}\textbf{S3} & \cellcolor{gray!30}23 & \cellcolor{gray!30}50 &  \cellcolor{gray!30}$320\times240$ & \cellcolor{gray!30}4.9636 & \cellcolor{gray!30}2.9874 & \cellcolor{gray!30}0.1294 & \cellcolor{gray!30}0.8168 & \cellcolor{gray!30}0.9580  & \cellcolor{gray!30}0.9892 \\
         & \textbf{S3} & 23 & 50 & $256\times192$ & 5.1478 & 3.1324 & 0.1337 & 0.8058 & 0.9542 & 0.9883 \\
         \hline\hline
        \multirow{6}*{FastDepth} & S0 & 23 & 0 & - & 6.1884 & 3.9174 & 0.1910 & 0.7147 & 0.9088 & 0.9684 \\\cline{2-11}
         & \textit{Synthetic} & 23 & 3 & $1940\times1080$ & 6.1257 & 3.8100 & 0.1895 & 0.7184 & 0.9182 & 0.9764 \\\cline{2-11}
         & S1 & 23 & 50 & $256\times192$ & 5.9277 & 3.6774 & 0.1854 & 0.7286 & 0.9240 & 0.9781 \\
         & S2 & 23 & 50 & $256\times192$ & 5.9417 & 3.6994 & 0.1884 & 0.7292 & 0.9223 & 0.9777 \\
          & \cellcolor{gray!30}\textbf{S3} & \cellcolor{gray!30}23 & \cellcolor{gray!30}50 &  \cellcolor{gray!30}$256\times192$ & \cellcolor{gray!30}5.6310 & \cellcolor{gray!30}3.5062 & \cellcolor{gray!30}0.1682 & \cellcolor{gray!30}0.7551 & \cellcolor{gray!30}0.9316 & \cellcolor{gray!30}0.9804 \\
         & \textbf{S3} & 23 & 50 & $320\times240$ & 5.8244 & 0.3613 & 0.1759 & 0.7374 & 0.9290 & 0.9792 \\
        \hline\hline
        \multirow{6}*{SPEED} & S0 & 23 & 0 & - & 5.3957 & 3.0473 & 0.1480 & 0.7797 & 0.9387 & 0.9841 \\\cline{2-11}
         & \textit{Synthetic} & 23 & 3 & $1940\times1080$ & 5.4219 & 3.1233 & 0.1565 & 0.7574 & 0.9307 & 0.9808 \\\cline{2-11}
         & S1 & 23 & 50 & $256\times192$ & 5.2321 & 2.9477 & 0.1409 & 0.7890 & 0.9445 & 0.9848 \\
         & S2 & 23 & 50 & $256\times192$ & 5.0945 & 2.8758 & 0.1401 & 0.7980 & 0.9476 & 0.9857 \\
         & \cellcolor{gray!30}\textbf{S3} & \cellcolor{gray!30}23 & \cellcolor{gray!30}50 &  \cellcolor{gray!30}$256\times192$ & \cellcolor{gray!30}4.9828 & \cellcolor{gray!30}2.8017 & \cellcolor{gray!30}0.1337 & \cellcolor{gray!30}0.8104 & \cellcolor{gray!30}0.9521  & \cellcolor{gray!30}0.9878 \\
         & \textbf{S3} & 23 & 50 & $320\times240$ & 5.2640 & 3.0663 & 0.1437 & 0.7823 & 0.9421 & 0.9839 \\
        \hline\hline
         \updated{\multirow{6}*{METER}} & \updated{S0} & \updated{23} &  \updated{0} & \updated{-} & \updated{4.8398} & \updated{2.7284} & \updated{0.1278} & \updated{0.8153} & \updated{0.9462} & \updated{0.9859} \\\cline{2-11}
         & \updated{\textit{Synthetic}} & \updated{23} &  \updated{3} & \updated{$1940\times1080$} & \updated{5.2139} & \updated{3.0725} & \updated{0.1468} & \updated{0.7753} & \updated{0.9428} & \updated{0.9847} \\\cline{2-11}
         & \updated{S1} & \updated{23} & \updated{50} & \updated{$256\times192$} & \updated{4.8961} & \updated{2.7206} & \updated{0.1275} & \updated{0.8118} & \updated{0.9512} & \updated{0.9864} \\
         & \updated{S2} & \updated{23} &  \updated{50} & \updated{$256\times192$} & \updated{4.7908} & \updated{2.8271} & \updated{0.1456} & \updated{0.7840} & \updated{0.9450} & \updated{0.9845} \\
         & \cellcolor{gray!30}\textbf{S3} & \cellcolor{gray!30}23 & \cellcolor{gray!30}50 & \cellcolor{gray!30}$256\times192$ & \cellcolor{gray!30}4.7288 & \cellcolor{gray!30}2.6833 & \cellcolor{gray!30}0.1308 & \cellcolor{gray!30}0.8155 & \cellcolor{gray!30}0.9533 & \cellcolor{gray!30}0.9875 \\
         & \updated{\textbf{S3}} & \updated{23} &  \updated{50} & \updated{$320\times240$} & \updated{4.7519} & \updated{2.6780} & \updated{0.1314} & \updated{0.8083} & \updated{0.9503} & \updated{0.9857} \\
         \hline
        \end{tabular}
    \end{center}
    \label{tab:kitti_test}
\end{table*}

\section{Experiments}
\label{sec:experiments}
In this section, we show the effectiveness of the proposed pipeline in terms of improvements obtained over the four chosen MDE models. 
The first performed analysis is computed with respect to indoor and outdoor D4D-generated datasets, i.e., when selected models are trained by adding the D4D-NYU and D4D-KITTI datasets. 
\upupdated{Subsequently, we investigate the effects of the different resolutions and amounts of RGBD data generated by D4D on the trained models.}
We conclude this section by analyzing the generalization performances on an unseen test dataset DIML/CVL RGB-D (DIML)\upupupupdated{, the estimation improvement over efficient variants of METER architecture} and with an analysis of similarity distances over probabilistic distributions.
We compare the obtained results with respect to S1, S2, S3, a baseline configuration (S0), i.e., when the models are trained on original datasets (NYU and KITTI), as well as an alternative augmentation schema based on synthetic datasets (\textit{Synthetic}).

\vspace{0.5em}
\noindent
\textbf{Indoor results.} 
The first analysis is performed on D4D-NYU dataset under different configurations (S\textit{i} with $i=[0,3]$ and \textit{Synthetic}), settings (NYU $=50K$ or NYU $=0$), number of generated samples \textit{(Add)} and D4D resolutions (\textit{Res}).
These training combinations have been taken in order to show how the presence of the original dataset and the generation resolution of the samples influence the estimation performances of chosen models.
\modified{
Precisely, we report the same tests over the four chosen reference MDE models with and without the original datasets (NYU), respectively in Table~\ref{tab:nyu_test_siNYU} and Table~\ref{tab:nyu_test_noNYU}, in order to understand differences,  similarities, and respective quantitative improvement obtained when using generated samples, i.e., how much D4D mimic original samples or how much those generated samples differs from original one.
Generally speaking, we noticed
}
that the proposed merging strategy (S3) has superior estimation performances in indoor scenarios with respect to all the compared configurations. 
Based on the achieved results, we derive that the closer the generation resolution of the samples is to the input resolution of the trained model, the better the estimation results, although the error difference is small (e.g., $2.2\%$ of the RMSE in the DenseDepth case).
This finding, based on the best D4D generation resolution, has been used in the experiments listed below and will be further investigated in the following ablation studies.
\updated{Moreover, we observe that by doubling the amount of generated data with respect to the original training dataset (from $50K$ to $100K$), the proposed configuration (S3) outperforms the baseline configuration (S0) and the \textit{Synthetic} datasets with an RMSE reduction equal to ($10.8\%$, $4.9\%$) on DenseDepth, ($14.7\%$, $9.7\%$) on FastDepth, ($11.6\%$, $11.1\%$) on SPEED and ($10.5\%$, $6.5\%$) on METER.
Furthermore, when trained only on D4D-NYU (NYU $=0$), S3 is able to achieve better performances than S0 in the case of FastDepth and SPEED, while slightly worse for DenseDepth and METER.} 
Contrarily, the synthetic RGBD data performs poorly without the original training dataset.
These results demonstrate the ability of D4D-generated samples to mimic real-world samples.
\updated{To summarize, the overall average percentage improvement obtained with the proposed training pipeline, computed with respect to the baseline configuration over the evaluation metrics used, is equal to $7.3\%$, $9.6\%$, $8.2\%$, and $6.2\%$ respectively for DenseDepth, FastDepth, SPEED, and METER.}

\modified{Finally, to have a complete understanding of the obtained improvement, we report in Figure~\ref{fig:densedepth_indoor} a qualitative comparison of the estimation performances of the DenseDepth model under the compared configurations, i.e., S\textit{i} with $i=[0, 3]$ and \textit{Synthetic}. 
Based on predicted depth maps and related difference maps\footnote{The difference map is computed as a per pixel-difference between predicted ($\hat{y}$) and expected ($y$) depth map.} reported for each configuration, we note that DenseDepth, in the synthetic configuration, produces the highest estimation error (more than $100cm$) with respect to compared setups.
Contrarily, S3 is the only configuration with an error range less than $80cm$ (demonstrated by darker difference map in Figure~\ref{fig:densedepth_indoor}).
Furthermore, we notice that all the compared predicted depth maps have well-defined contours.
However, in the reported case, the proposed configuration (S3) is able to correctly estimate distances in the situation where all the others fail, i.e., where the scene distance varies rapidly (e.g., behind a wall); we highlight this area on the difference map with a dashed red rectangle.}

\begin{figure*}[t]
    \centering
    \includegraphics[width=\linewidth]{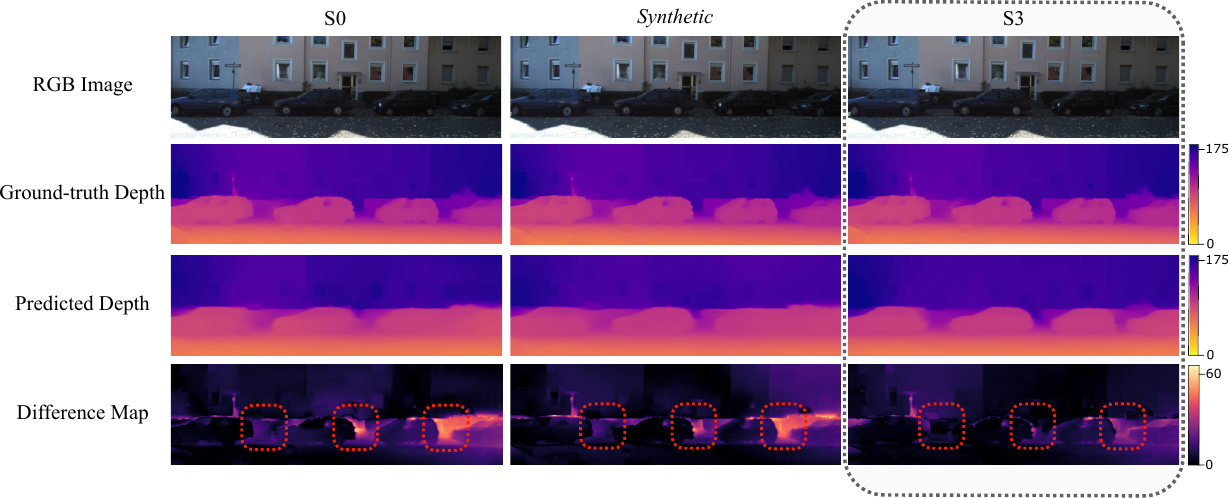} 
    \caption{\modified{\textbf{Outdoor results.} Qualitative analysis of the estimated prediction obtained with DenseDepth method. The model has been tested on KITTI (outdoor) dataset. S0 is the baseline setup, i.e., when DenseDepth is trained only on KITTI dataset. In \textit{Synthetic} setup, the model has been trained over KITTI and SYNTHIA-SF datasets. In the proposed configuration (S3), the model has been trained over KITTI and $50K$ samples taken from our proposed D4D-KITTI datasets generated at a resolution of $320\times240$. The Difference Map is computed as a per pixel-difference between predicted ($\hat{y}$) and expected depth ($y$), while the reported colorbars are used to emphasize the depth/error range in decimeters ($dm$).}}
    \label{fig:densedepth_outdoor}
\end{figure*}

\vspace{0.5em}
\noindent
\textbf{Outdoor results.} Along with the previous findings, the proposed method (S3) achieves notable estimation improvements also in the outdoor scenario, especially when the D4D generation resolution is close to the MDE model input resolution.
We report in Table~\ref{tab:kitti_test} the results obtained by the selected MDE models when trained on KITTI dataset and in combination with D4D-KITTI or the synthetic SYNTHIA-SF dataset.
\updated{Precisely, the maximum RMSE reduction with respect to S0 and the \textit{Synthetic} dataset is obtained by tripling the amount of training data, and it is equal to ($4.7\%$, $6.3\%$) on DenseDepth, ($9.1\%$, $8.1\%$) on FastDepth, ($8.3\%$, $8.8\%$) on SPEED, and ($2.3\%$, $9.3\%$) on METER.}
However, we cannot rule out that further improvements could be obtained by greatly increasing the number of generated samples.
\updated{Summarizing, the overall average percentage improvement achieved with the proposed training pipeline, when compared with S0, is equal to $4.0\%$, $6.7\%$, $5.7\%$, and $\simeq1.0\%$ respectively, for DenseDepth, FastDepth, SPEED, and METER.
The latter results obtained for the hViT architecture are most likely attributed to the D4D generation resolution. 
Consequently, similar to the indoor scenario, we expect comparable RMSE reductions to the CNN architectures in the case of images generated at the same working resolution of METER.}
These results confirm the soundness of D4D for increasing the performances of any kind of MDE model.
\modified{
Finally, we report in Figure~\ref{fig:densedepth_outdoor} a qualitative comparison
for the estimation performances of the DenseDepth model in S0, \textit{Synthetic}, and S3 configurations.
Based on the reported predictions and associated difference maps, we noticed that the maximum depth error for all the configurations is in between $(50, 60) dm$. 
However, the proposed setup (S3) predicts object edges and overall distances more precisely than the other configurations; we highlight these areas on the difference map with three dashed red circles (the darker is the area the better).
}

\begin{table*}[!ht]
    \footnotesize
    \centering
    \caption{Generalization performances of DenseDepth on DIML/CVL RGB-D test dataset. The proposed strategy is in bold, while the optimal configuration is highlighted in gray.}
    \begin{center}
        \begin{tabular}{ c | c | c c c | c c c | c c c }
        \hline
        Model & Configuration & NYU [K] & \textit{Add} [K] & \textit{Res} [pix] & \cellcolor{orange!20}RMSE$\downarrow$ [m] & \cellcolor{orange!20}MAE$\downarrow$ [m]  & \cellcolor{orange!20}Abs$_{Rel}$$\downarrow$ & \cellcolor{blue!15}$\delta_1$$\uparrow$ & \cellcolor{blue!15}$\delta_2$$\uparrow$ & \cellcolor{blue!15}$\delta_3$$\uparrow$ \\
        \hline\hline
        \multirow{9}*{DenseDepth} & S0 & 50 & 0 & - & 0.8723 & 0.7295 & 0.1268 & 0.4466 & 0.7968 & 0.9337 \\\cline{2-11}
         & \textit{Synthetic} & 0 & 50 & $320\times240$ & 1.0901 & 0.8999 & 0.3738 & 0.4221 & 0.7188 & 0.8800 \\
         & \textit{Synthetic} & 0 & 150 & $320\times240$ & 1.0510 & 0.8721 & 0.3747 & 0.4294 & 0.7248 & 0.8766 \\\cline{2-11}
         & S1 & 0 & 50  & $320\times240$ & 0.8443 & 0.7126 & 0.2696 & 0.4876 & 0.8225 & 0.9331 \\
         & S2 & 0 & 50 & $320\times240$ & 0.9417 & 0.7975 & 0.1432 & 0.4005 & 0.7255 & 0.8943 \\
         & \cellcolor{gray!30}\textbf{S3}  & \cellcolor{gray!30}0 & \cellcolor{gray!30}50 & \cellcolor{gray!30}$320\times240$ & \cellcolor{gray!30}0.7959 & \cellcolor{gray!30}0.6660 & \cellcolor{gray!30}0.2486 & \cellcolor{gray!30}0.5069 & \cellcolor{gray!30}0.8381 & \cellcolor{gray!30}0.9540 \\
         & \textbf{S3} & 0 & 50 & $256\times192$ & 0.8142 & 0.6864 & 0.2730 & 0.4998 & 0.8278 & 0.9365 \\\cline{2-11}
         & \updated{\textbf{S3}} & \updated{0} & \updated{100} & \updated{$320\times240$} & \updated{0.8001} & \updated{0.6701} & \updated{0.2522} & \updated{0.4921} & \updated{0.8377} & \updated{0.9537}\\
         & \updated{\textbf{S3}} & \updated{0} & \updated{150} & \updated{$320\times240$} & \updated{0.7914} & \updated{0.6623} & \updated{0.2439} & \updated{0.5116} & \updated{0.8421} & \updated{0.9548} \\
        \hline
        \end{tabular}
    \end{center}
    \label{tab:diml_blind}
\end{table*}

\begin{figure*}[t]
    \centering
    \includegraphics[width=\linewidth]{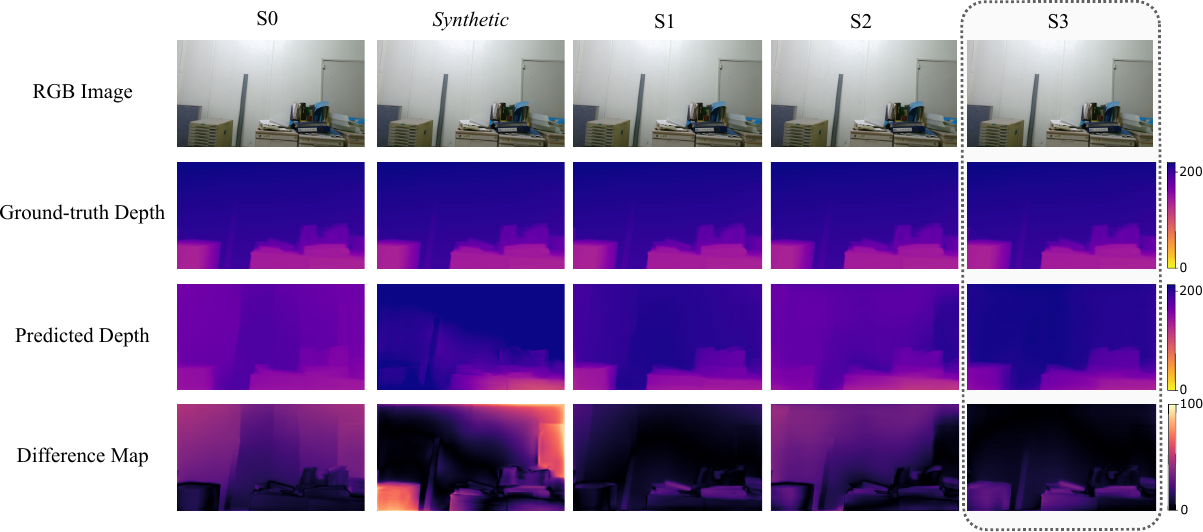} 
    \caption{\modified{\textbf{Generalization.} 
    Qualitative analysis of the estimated prediction obtained with DenseDepth method. The model has been tested in blind condition (i.e., without fine-tuning) on DIML/CVL RGB-D dataset
    when trained on a different indoor dataset, i.e., NYU for S0, SceneNet for Synthetic, and D4D-NYU for S1, S2, and S3.
    The Difference Map is computed as a per pixel-difference between predicted ($\hat{y}$) and expected depth ($y$), while the reported colorbars are used to emphasize the depth/error range in centimeters ($cm$).}}
    \label{fig:densedepth_diml_generalization}
\end{figure*}

\vspace{0.5em}
\noindent
\textbf{Generalization.} 
After showing the efficacy of the proposed solution in the two most common MDE scenarios, we illustrate the generalization performances of DenseDepth in a blind test, i.e., when the model is trained and tested over two different datasets without fine-tuning. 
In detail, we used the selected model as in previous indoor analysis and tested it on a different real-world dataset (DIML). 
We report the obtained results in Table~\ref{tab:diml_blind}.
It is possible to point out that when the model is trained on S3 configuration\updated{, with the same amount of training samples ($Add=50K$)}, it outperforms the generalization performances of S0 (NYU). 
In the case of \textit{Synthetic} (SceneNet), such behavior is evident even when the number of training samples is increased to $150K$. 
Moreover, using $320\times240$ pixels as D4D generation resolution, S3 achieves over S0 and \textit{Synthetic} data an RMSE reduction equal to ($8.7\%$, $26.9\%$) respectively.
\updated{Furthermore, the increase ($100K$ and $150K$) of D4D-generated samples results in comparable estimation performances with the previously analyzed S3 configuration ($Add=50K$), as shown in Table~\ref{tab:diml_blind}, which does not justify the time required to produce the additional samples.}
\modified{
More in detail, Figure~\ref{fig:densedepth_diml_generalization} reports a qualitative analysis of DenseDepth model trained on NYU, SceneNet, or D4D-NYU (separately) and tested (without fine-tuning) on the DIML/CVL RGB-D dataset over the compared configurations of Table~\ref{tab:diml_blind}.
Based on predicted depth maps and related difference maps for each configuration, it is possible to notice that S3 achieves a lower estimation error than all the other configurations. 
Precisely, with a maximum distance error of almost $40cm$ with respect to the $100cm$ and $57cm$ achieved by synthetic and baseline (S0) setups.
}
These quantitative and qualitative comparisons demonstrate the superior performances of the proposed D4D-NYU dataset even when testing \modified{MDE models} on an unseen dataset.

\begin{figure*}[t]
    \centering
    \includegraphics[width=\linewidth]{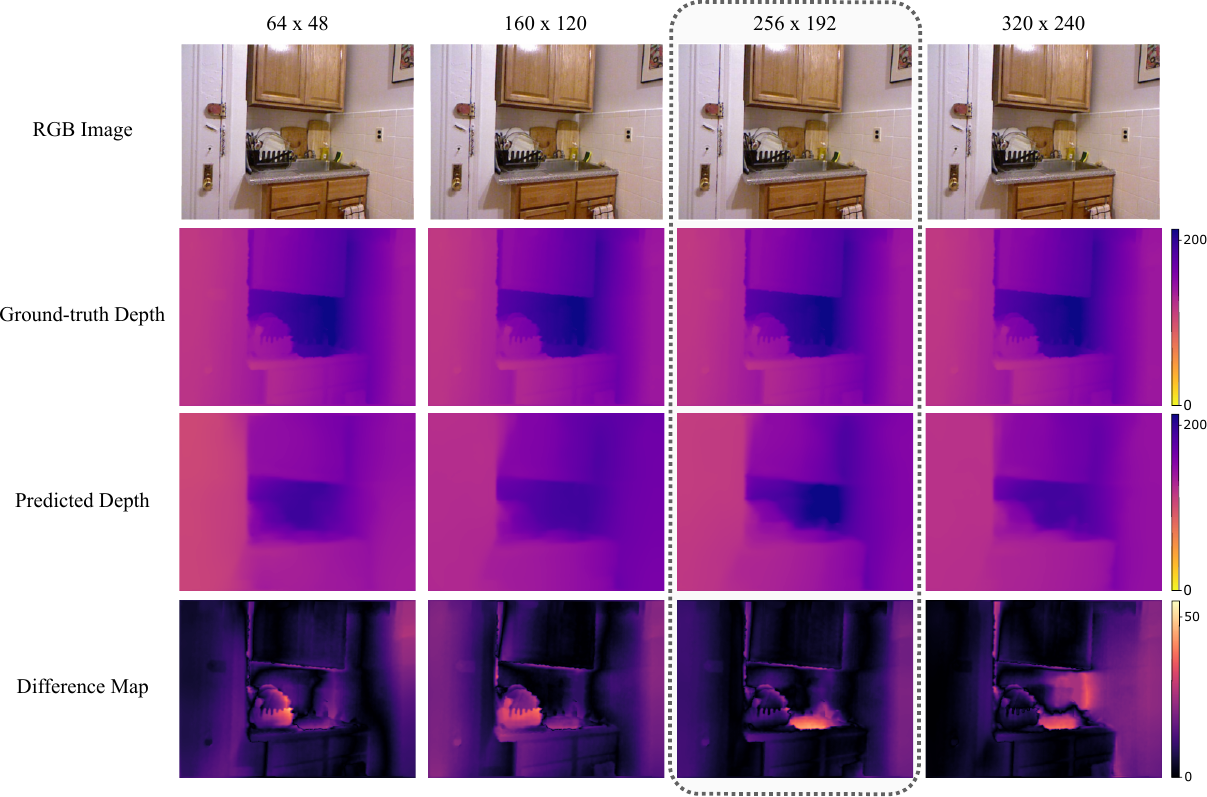} 
    \caption{\modified{\textbf{Image resolution.} Qualitative analysis of the estimated prediction obtained with FastDepth method. The model has been tested on NYU (indoor) dataset and trained under S3 settings over NYU and D4D-NYU datasets, where its samples have been generated at different resolutions.
    The Difference Map is computed as a per pixel-difference between predicted ($\hat{y}$) and expected depth ($y$), while the reported colorbars are used to emphasize the depth/error range in centimeters ($cm$).}}
    \label{fig:fastdepth_resolution}
\end{figure*}

\vspace{0.5em}
\noindent
\textbf{Image resolution.} 
In previous experiments, we showed that the image resolution of D4D-generated samples leads to better depth estimation performances when it is closer to the input image resolution of the trained model.  
Therefore, we report in Table~\ref{tab:resolutions} a detailed analysis of the effects of D4D-generated resolutions over deep (DenseDepth) and shallow (FastDepth) MDE models. 
This experiment has been performed on indoor samples (D4D-NYU) with the best parameters' setup, i.e., S3 configuration, NYU $=50K$, and $Add=100K$. 
The previous trend is confirmed since working with a generation resolution significantly different from the model input leads to a noticeable performance decrease, with a maximum difference on the RMSE equal to ($19.9\%$, $15.3\%$) and an overall averaged percentage reduction of ($17.3\%$, $12.2\%$) on DenseDepth and FastDepth. 
Thanks to this fact, we could keep limited computational requirements needed to generate RGBD samples, avoiding the use of unnecessary high resolutions.
\begin{table}[t]
    \footnotesize
    \centering
    \caption{Quantitative comparison of MDE models trained on subsets of D4D-NYU generated at different resolutions. The optimal values for each compared model are highlighted in gray.}
    \begin{center}
        \begin{tabular}{ c | c | c c c }
        \hline
        Model & \textit{Res} [pix] & \cellcolor{orange!20}RMSE$\downarrow$ [m] & \cellcolor{orange!20}Abs$_{Rel}$$\downarrow$ & \cellcolor{blue!15}$\delta_1$$\uparrow$ \\
        \hline\hline
        \multirow{4}*{DenseDepth} & $64\times48$ & 0.5595 & 0.1595 & 0.7678\\
         & $160\times120$ & 0.4829 & 0.1342 & 0.8258 \\
         & $256\times192$ & 0.4578 & 0.1286 & 0.8376 \\
         & \cellcolor{gray!30}$320\times240$ & \cellcolor{gray!30}0.4480 & \cellcolor{gray!30}0.1236 & \cellcolor{gray!30}0.8499 \\
        \hline\hline
        \multirow{4}*{FastDepth} & $64\times48$ & 0.5880 & 0.1758 & 0.7410\\
         & $160\times120$ & 0.5443 & 0.1616 & 0.7816 \\
         & \cellcolor{gray!30}$256\times192$ & \cellcolor{gray!30}0.4980 & \cellcolor{gray!30}0.1414 & \cellcolor{gray!30}0.8119 \\
         & $320\times240$ & 0.5103 & 0.1492 & 0.7903 \\
         \hline
        \end{tabular}
    \end{center}
    \label{tab:resolutions}
\end{table}
\modified{
Finally, Figure~\ref{fig:fastdepth_resolution} reports a qualitative analysis of FastDepth architecture (other models show similar behavior) when trained on NYU and the proposed D4D-NYU dataset (S3 settings) when its samples are generated at different resolutions ranging from $64\times48$ to $320\times240$ pixels. 
Based on predicted depth maps and related difference maps for each generation resolution, we qualitatively confirm the fact that the closer the generation resolution of D4D to the input resolution of FastDepth, the better is the estimation for the MDE model. 
In fact, as noticed, the dataset generated at an image resolution of $256\times192$ pixels, which is closer to FastDepth's input resolution ($224\times224$), has a lower error distribution. This can be noticed from the dark region areas that are larger with respect to the other predictions (underlined by the gray dashed rectangle in Figure~\ref{fig:fastdepth_resolution}).
}

\updated{
Based on the obtained findings, we assume that the just described behavior, due to the different D4D generation resolutions, is caused by the varying feature extraction capabilities of each MDE architecture.
More in detail, since each MDE architecture has been developed to work with a specific input resolution, it follows that this parameter defines the quantity of information (pixels) that the model is able to process in order to ensure optimal performance.
Consequently, the closer the resolution used to generate samples is to the network's working resolution, the better the performance; in contrast, samples that are larger/smaller than the working resolution of the network will be compressed/expanded, thus resulting in information loss or inaccurate data.
}

\vspace{0.5em}
\noindent
\upupdated{
\textbf{Amount of generated samples.} Once the optimal generation resolution has been analyzed, in this ablation study, we investigate how different amounts of generated samples impact the performance of MDE models.
More in detail, we study the behavior of FastDepth architecture when the number of D4D-generated (training) samples varies; precisely, we examine a data range between 0 and 250K RGBD samples generated by D4D-NYU in the optimal S3 configuration at the resolution of 256 x 192 pixels.
We report the obtained results in the two compared setups, i.e., with and without the original training dataset (NYU), in Table~\ref{tab:amount_of_added_samples}. 

Based on the obtained results, it can be noticed that the higher estimation performances are obtained with the addition of $200K$ generated training data ($Add=200K$). 
More in detail, we obtain an average RMSE reduction of $7.9\%$ and $4.6\%$ when the best performing model is compared with respect to the other configurations ($Add=i*50K$ with $i\in[0,3]$) in the two analyzed scenarios, i.e., when the original training dataset is used (NYU=$50K$) and when it is not considered (NYU=$0K$).
\begin{table}[t]
    \footnotesize
    \centering
    \caption{\upupdated{Quantitative comparison of FastDepth model trained on different amount of D4D-NYU generated samples (S3 configuration) at the resolution of $256 \times 192$ pixels. The best values for each compared setup are highlighted in gray.}}
    \resizebox{\linewidth}{!}{%
    \begin{tabular}{ c | c c | c c c }
        \hline
        Model & NYU [K] & Add [K] & \cellcolor{orange!20}RMSE$\downarrow$ [m] & \cellcolor{orange!20}Abs$_{Rel}$$\downarrow$ & \cellcolor{blue!15}$\delta_1$$\uparrow$ \\
        \hline\hline
        \multirow{12}*{FastDepth} & 50 & 0 & 0.5714 & 0.1751 & 0.7535\\\cline{2-6}
        & 50 & 50 & 0.5585 & 0.1643 & 0.7666 \\
        & 50 & 100 & 0.4980 & 0.1414 & 0.8119 \\
        & 50 & 150 & 0.4962 & 0.1411 & 0.8121 \\
        & \cellcolor{gray!30}50 & \cellcolor{gray!30}200 & \cellcolor{gray!30}0.4919 & \cellcolor{gray!30}0.1406 & \cellcolor{gray!30}0.8127 \\
        & 50 & 250 & 0.5076 & 0.1517 & 0.7981 \\ \cline{2-6}
        & 0 & 0 & - & - & - \\
        & 0 & 50 & 0.5996 & 0.1746 & 0.7500 \\
        & 0 & 100 & 0.5502 & 0.1730 & 0.7649 \\
        & 0 & 150 & 0.5449 & 0.1619 & 0.7665 \\
        & \cellcolor{gray!30}0 & \cellcolor{gray!30}200 & \cellcolor{gray!30}0.5397 & \cellcolor{gray!30}0.1607 & \cellcolor{gray!30}0.7678 \\
        & 0 & 250 & 0.5444 & 0.1616 & 0.7629 \\
         \hline
    \end{tabular}%
    }
    \label{tab:amount_of_added_samples}
\end{table}
Based on the two compared configurations, we can note that when comparing the best-performing setup with the best one ($Add=100K$) reported in Table~\ref{tab:nyu_test_siNYU} and Table~\ref{tab:nyu_test_noNYU} (also reported in Table~\ref{tab:amount_of_added_samples}), the RMSE reduction is limited to $1.2\%$ and $1.9\%$, respectively.
Moreover, when compared to the $Add=250K$ setups, the $Add=200K$ ones results in an RMSE reduction of $3.1\%$ (NYU$=50$) and $0.9\%$ (NYU$=0$).
Consequently, we can assume that the $Add=200K$ setup is FastDepth's best configuration with respect to the amount of generated samples; however, when considering the time required to generate a larger number of RGBD data and the limited percentage improvement, we can conclude that $100K$ samples are a good trade-off, ensuring good estimation performance on the NYU dataset while limiting the overall computational time. 
}

\vspace{0.5em}
\noindent
\upupupupdated{\textbf{Additional experiments on efficient ViT.}
Once the main parameters of D4D have been analyzed, we present some additional results on efficient ViT architectures to emphasize the proposed solution's versatility. 
We outline the following analysis motivated by the practical applicability of MDE models on embedded/mobile devices, which are usually characterized by limited computational powers.
In order to infer on such devices, factors like reduced network computational capabilities, number of trainable parameters, or model depth typically result in a reduction of the estimation performances.
Consequently, this analysis investigates the percentage boost that D4D is able to achieve when combined with efficient architectures.
In particular, we analyze the performance improvement of the proposed pipeline across two efficient METER configurations, namely, Meta-METER (MetaM) and Pyra-METER (PyraM) proposed in~\cite{effmeter}. 
The latter architectures were developed by exploiting the efficiency capabilities of MetaFormer~\cite{yu2022metaformer} and Pyramid Vision Transformer~\cite{wang2021pyramid}, which aims to reduce/linearize the computational cost of self-attention.  

We compare the reported architectures using the same METER's optimal\footnote{For the NYU dataset: configuration S3, \textit{Add$=100K$}, \textit{Res. $256\times192$}. For the KITTI dataset: configuration S3, \textit{Add$=50K$}, \textit{Res. $256\times192$}.} hyperparameters identified in Table\ref{tab:nyu_test_siNYU} and Table\ref{tab:kitti_test} respectively for the NYU and KITTI datasets. 
Based on the obtained results (Table \ref{tab:efficientViT}), we can note an average percentage RMSE reduction of $6.4\%$ and $\delta_1$ increment of $2.0\%$ when the D4D pipeline is used instead of a standard training pipeline.
As a result, it can be noticed that in this scenario, where model learning capabilities are limited with respect to deeper architectures due to computational constraints introduced by embedded devices, the proposed pipeline still provides a good percentage boost for the model's estimation performances.
}

\begin{table}[h]
    \footnotesize
    \centering
    \caption{\upupupupdated{Quantitative comparison across efficient hViT configurations. The \cmark and \xmark are used to indicate when D4D-generated data are employed.}}
    \begin{center}
        \begin{tabular}{ l | c | c c | c c }
             \hline
             \multirow{2}*{\upupupupdated{Model}} & \multirow{2}*{\upupupupdated{D4D}} &\multicolumn{2}{c|}{\upupupupdated{NYU}} & \multicolumn{2}{c}{\upupupupdated{KITTI}} \\
             & & \cellcolor{orange!20}\upupupupdated{RMSE$\downarrow$ [m]} & \cellcolor{blue!15}\upupupupdated{$\delta_1$$\uparrow$} & \cellcolor{orange!20}\upupupupdated{RMSE$\downarrow$ [m]} &  \cellcolor{blue!15}\upupupupdated{$\delta_1$$\uparrow$} \\
             \hline
             \hline
             \upupupupdated{\multirow{2}*{METER}} & \upupupupdated{\xmark} & \upupupupdated{0.5112} & \upupupupdated{0.8138} & \upupupupdated{4.8398} & \upupupupdated{0.8153} \\
              & \cellcolor{gray!30}\upupupupdated{\cmark} & \cellcolor{gray!30}\upupupupdated{0.4574} & \cellcolor{gray!30}\upupupupdated{0.8357} & \cellcolor{gray!30}\upupupupdated{4.7288} & \cellcolor{gray!30}\upupupupdated{0.8155} \\
             \hline
             \hline
             \upupupupdated{\multirow{2}*{MetaM}} & \upupupupdated{\xmark} & \upupupupdated{0.5058} & \upupupupdated{0.8111} & \upupupupdated{4.9493} & \upupupupdated{0.8014} \\
              & \cellcolor{gray!30}\upupupupdated{\cmark} & \cellcolor{gray!30}\upupupupdated{0.4556} & \cellcolor{gray!30}\upupupupdated{0.8373} & \cellcolor{gray!30}\upupupupdated{4.6714} & \cellcolor{gray!30}\upupupupdated{0.8166} \\
             \hline
             \hline
             \upupupupdated{\multirow{2}*{PyraM}} & \upupupupdated{\xmark} & \upupupupdated{0.5196} & \upupupupdated{0.8062} & \upupupupdated{5.2308} & \upupupupdated{0.7652} \\
              & \cellcolor{gray!30}\upupupupdated{\cmark} & \cellcolor{gray!30}\upupupupdated{0.4944} & \cellcolor{gray!30}\upupupupdated{0.8139} & \cellcolor{gray!30}\upupupupdated{4.9652} & \cellcolor{gray!30}\upupupupdated{0.7737} \\
              \hline
        \end{tabular}
    \end{center}
    \label{tab:efficientViT}
\end{table}

\vspace{0.5em}
\noindent
\textbf{Analysis on feature space.} We conclude the result section by performing similarity measurements among different configurations on the feature space in order to provide an in-depth explanation of the obtained results.
More in detail, we analyze the learning capabilities of D4D configurations (S1, S2, and S3) with respect to the NYU training setup (S0).
We extract the visual features characterizing each dataset with two pretrained neural networks (initialized on ImageNet): the ResNet18~\cite{resnet} and the EfficienNetB4~\cite{effnet}.
This procedure is performed by removing the last classification layer (fully connected) from each respective model.
Therefore, a final embedding vector of each dataset is obtained as the average features vector extracted from $50K$ input samples.
Subsequently, we compute the distance between the mean of the embedding vectors using two evaluation metrics: the Euclidean distance (ED) and the Hillinger distance (HD)~\cite{pollard2002user}. 
Table~\ref{tab:features_test} shows such differences computed between the embedding vectors related to each configuration and the NYU test dataset.
\begin{table}[h]
    \footnotesize
    \centering
    \caption{Embedding vectors' distances computed between each configurations (S\textit{i} with $i=[0,3]$) and NYU test set. Each subset counts $50K$ training samples.}
    \begin{center}
        \begin{tabular}{ l | c | c c | c c }
             \hline
             \multirow{2}*{Model} & \multirow{2}*{Configuration} & \multicolumn{2}{c|}{RGB} & \multicolumn{2}{c}{Depth} \\
             & & ED & HD & ED & HD \\
             \hline\hline
             \multirow{4}*{ResNet18} & S0 & 0.1158 & 0.0771 & 0.1243 & 0.0826 \\ 
              & S1 & 0.2626 & 0.1720 & 0.2739 & 0.1815 \\ 
              & S2 & 0.2384 & 0.1597 & 0.2568 & 0.1719 \\
              & \cellcolor{gray!30}\textbf{S3} & \cellcolor{gray!30}0.3636 & \cellcolor{gray!30}0.2403 & \cellcolor{gray!30}0.3608 & \cellcolor{gray!30}0.2361 \\
              \hline
              \hline
              \multirow{4}*{EffB4} & S0 & 0.7714 & 0.4442 & 1.2708 & 0.7262 \\ 
              & S1 & 1.7724 & 1.0307 & 2.0593 & 1.1886 \\ 
              & S2 & 1.4572 & 0.8330 & 1.8667 & 1.0584 \\
              & \cellcolor{gray!30}\textbf{S3} & \cellcolor{gray!30}1.9611 & \cellcolor{gray!30}1.1356 & \cellcolor{gray!30}2.1222 & \cellcolor{gray!30}1.2283 \\
              \hline
        \end{tabular}
    \end{center}
    \label{tab:features_test}
\end{table}

Based on reported values, S3 has higher values both for ED and HD rather than other configurations.
Moreover, observing the metrics reported in  Table~\ref{tab:nyu_test_siNYU}, Table~\ref{tab:nyu_test_noNYU}, and Table~\ref{tab:diml_blind} we noticed that the increasing distances correspond to greater estimation performances.
Therefore, \modified{
without loss of generality,
} we derive 
that the higher the distance of the features from the test dataset, the better the performance of the MDE model.
We hypothesize that a greater distance corresponds to stronger generalization capabilities due to a more efficient covering of heterogeneous samples.

\section{Conclusion}
\label{sec:conclusions}
This paper presents a novel training pipeline composed of D4D, a custom 4-channels DDPM to produce realistic RGBD samples used to improve the estimation performances of deep and shallow MDE models. 
\updated{The proposed methodology demonstrates superior performances with respect to synthetically generated datasets in indoor and outdoor scenarios, with an average RMSE reduction equal to $8.2\%$ and $8.1\%$.
Moreover, our solution achieves an RMSE reduction equal to 
$11.9\%$ and $6.1\%$  with respect to the baseline indoor NYU Depth v2 and outdoor KITTI datasets.}
We hope that our method, together with the generated datasets (D4D-NYU and D4D-KITTI), will encourage the combined use of DDPM with deep learning architectures to address the lack of labeled training data in a variety of computer vision applications.
A key element of the proposed strategy is the use of real-world images to generate novel augmented samples, thus improving the estimation and generalization of MDE model capabilities for deploying in real-case scenarios.

Our technique is applied to tackle the MDE task, where the generated depth map is crucial to obtain accurate performance. 
However, the generated RGBD samples could also contribute to other applications, such as monocular SLAM or other computer vision tasks where a fourth (depth) channel can be used to improve standard RGB approaches, as in semantic segmentation~\cite{cheng2017locality}, human action recognition \cite{9841515} and object detection \cite{9789193}.
\upupdated{Consequently, in the future, we will further evaluate our method and employ generated samples in different RGBD tasks, study their performances on different architectures, and propose new diffusion architectures specifically tailored for depth data.}

\section*{Acknowledgments}
This study has been partially supported by the Italian Ministry of Enterprises and Made in Italy (Ministero delle Imprese e del Made in Italy - MIMIT) with the project PMDI 2023-2026, Sapienza University of Rome project 2022–2024 “EV2” (003\_009\_22), and project 2022–2023 “RobFastMDE”.

\bibliographystyle{IEEEtran}
\bibliography{main}

\begin{IEEEbiography}[{\includegraphics[width=1in,height=1.25in,clip,keepaspectratio]{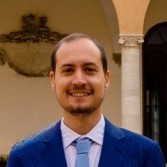}}]{LORENZO PAPA} is a Ph.D. student in Computer Science Engineering. 
He collaborates with AlcorLab at the Department of Computer, Control, and Management Engineering, Sapienza University of Rome, Italy.
He is a Visiting Researcher at the School of Electrical and Information Engineering, Faculty of Engineering and Information Technology, The University of Sydney, Australia. 
He received the B.S. degree in Computer and Automation Engineering and the M.S. degree in Artificial Intelligence and Robotics from Sapienza University of Rome, Italy, in 2019 and 2021, respectively.
His main research interests are Deep Learning, Computer Vision, and Cyber Security.
\end{IEEEbiography}

\begin{IEEEbiography}[{\includegraphics[width=1in,height=1.25in,clip,keepaspectratio]{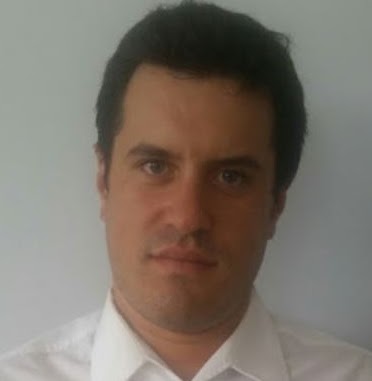}}]{PAOLO RUSSO} is an Assistant Researcher at AlcorLab in DIAG department, University of Rome Sapienza, Italy. 
He received the B.S. degree in Telecommunication Engineering from Università degli studi di Cassino, Italy, in 2008, and the M.S. degree in Artificial Intelligence and Robotics from University of Rome La Sapienza, Italy, in 2016. 
He received Ph.D. degree in Computer Science from University of Rome La Sapienza in 2020.
From 2018 to 2019, he has been a researcher at Italian Institute of Technology (IIT) in Tourin, Italy.
His main research interests are Deep Learning, Computer Vision, Generative Adversarial Networks, and Reinforcement Learning.
\end{IEEEbiography}

\begin{IEEEbiography}[{\includegraphics[width=1in,height=1.25in,clip,keepaspectratio]{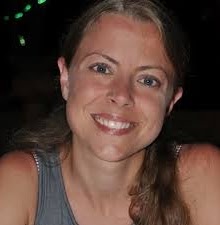}}]{IRENE AMERINI} (M’17) received Ph.D. degree in computer engineering, multimedia, and telecommunication
from the University of Florence, Italy, in 2010. She is currently Associate Professor with the Department of
Computer, Control, and Management Engineering A. Ruberti, Sapienza University of Rome, Italy. Her main
research activities include digital image processing, computer vision and multimedia forensics. She is a
member of the IEEE Information Forensics and Security Technical Committee, the EURASIP TAC Biometrics, Data Forensics, and Security, and the IAPR TC6 - Computational Forensics Committee.
\end{IEEEbiography}

\vfill

\end{document}